# Quantitative Method for Security Situation of the Power Information Network Based on the Evolutionary Neural Network

Quande Yuan[1], 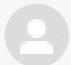 Yuzhen Pi[1,2], 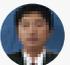 Lei Kou[3]*, 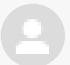 Fangfang Zhang[3] and 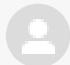 Bo Ye[4]

[1] Changchun Institute of Technology, Changchun, China
[2] National Local Joint Engineering Research Center for Smart Distribution Grid Measurement and Control with Safety Operation Technology, Changchun Institute of Technology, Changchun, China
[3] Qilu University of Technology (Shandong Academy of Sciences), Qingdao, China
[4] State Grid Xinjiang Electric Power Research Institute, Urumqi, China

Cybersecurity is the security cornerstone of digital transformation of the power grid and construction of new power systems. The traditional network security situation quantification method only analyzes from the perspective of network performance, ignoring the impact of various power application services on the security situation, so the quantification results cannot fully reflect the power information network risk state. This study proposes a method for quantifying security situation of the power information network based on the evolutionary neural network. First, the security posture system architecture is designed by analyzing the business characteristics of power information network applications. Second, combining the importance of power application business, the spatial element index system of coupled interconnection is established from three dimensions of network reliability, threat, and vulnerability. Then, the BP neural network optimized by the genetic evolutionary algorithm is incorporated into the element index calculation process, and the quantitative model of security posture of the power information network based on the evolutionary neural network is constructed. Finally, a simulation experiment environment is built according to a power sector network topology, and the effectiveness and robustness of the method proposed in the study are verified.

## Introduction

Along with the deep transformation of the new power grid, China's power network has gradually established a set of computer network architecture from the national grid data center down to the provinces and municipalities (Dileep, 2020; Huang et al., 2021). However, while the new form of the power grid is more intelligent, a large number of power electronic devices and digital monitoring terminals are connected. The security risk brought by the information side of the power grid has become one of the important hidden dangers that threaten the security of the power grid. (Pliatsios et al., 2020; Butt et al., 2021). Existing network attack methods are more threatening, purposeful, and large-scale. The full realization of the security situation awareness of the power information network can better assist the security managers to grasp the power grid status in time and maintain the power grid security (Lei et al., 2021). Therefore, in view of the development trend and characteristics of the new power information network, it is of great practical significance to study practical and effective methods for quantifying the security posture (Wang and Govindarasu, 2020).

In recent years, scholars around the world have considered using different methods to accurately quantify the security situation of power information networks. Wang et al. (2016) summarize the functions of various network protection systems and technologies in security defense by analyzing the impact of the structure of the communication network of electric power companies on network security. It proposed a specific scheme for applying firewall and the intrusion detection system to the communication network of electric power companies and discussed the construction of related technologies and the network security system (Li et al., 2022). Based on the system security engineering capability maturity model (SSE-CMM) proposed in Li B. et al. (2020) and Ganjkhani et al. (2021), the comprehensive fuzzy judgment method of power information security engineering is used to establish the power information network model, and the entropy weight coefficient of each evaluation factor is determined using the entropy weight coefficient method. The validity of the established model is verified through actual arithmetic cases. Panda and Das (2021) establish a new protection model for power communication networks and further clarifies the multiple elements that interfere with the security problems of power communication networks. Based on the fuzzy mathematical theory, Li Q. et al. (2020) and Qu et al. (2021) calculate element weight coefficients by constructing a fuzzy evaluation model for multiple elements, and then measure the security posture and network defense performance of the power communication network.

In summary, domestic research on power information network security is limited to traditional internet security awareness techniques or risk assessment methods based on flat systems. The quantitative index system of security posture lacks the combination with the application business of the power information network, which leads to the quantitative results cannot monitor the power information network in all aspects.

In this study, integrating business importance and network performance, we proposed a method to quantify the security posture of the power information network based on the evolutionary neural network and realized the comprehensive quantitative analysis of power information network posture. The main contributions of this study are as follows:

1) By considering the business characteristics of power grid applications, a new type of network security situational awareness architecture is designed.
2) A three-dimensional element index system with multi-dimensional network attributes is established to ensure the coupling and interconnection of element indexes and power application business, at the same time, the mathematical characterization of key element indexes is carried out.
3) The back propagation (BP) neural network optimized by the genetic evolutionary algorithm is incorporated into the calculation process of elemental indicators. The quantitative model of the security posture of the power information network is constructed, which effectively realizes the efficient calculation of the comprehensive perception process of the network security posture and the accurate quantification of the results.

This article is organized as follows: Section 2 provides a brief description of the architecture of the security posture quantification system for power information networks. The method for quantifying the security posture of power information networks is described in detail in section 3. Section 4 verifies the effectiveness of the proposed method based on the simulated experimental environment. Finally, conclusions and future recommendations are presented in section 5.

## Power Information Network Security Posture Quantification System Architecture

The power information network includes a multi-layer network structure of power grid data centers, backbone networks, research networks, office networks, and provincial and local municipal networks (Liu J. et al., 2020). This means that the security situation must be a combination of security situations of each network in a certain way (Cao et al., 2019). At the same time, different departments and sub-networks have deployed different firewalls, intrusion detection systems, security audit devices, and security hierarchies. If network security data information cannot be collected from sub-networks and the information is processed centrally and interactively, the network data information is isolated from each other, leading to the emergence of security network security silos. Therefore, an accurate and intuitive situation value is an indispensable part of the security situation awareness of the power information network (Wang et al., 2021; Zhang et al., 2021).

The network security situational awareness architecture is the basis of situational awareness research for power private communication networks (Liang et al., 2021), and whether its design is reasonable to directly affect the overall feasibility of network situational awareness. In the article, the network security situation quantification of the power information network (PIN-NSSQ) architecture is designed by analyzing three levels: network hierarchy, data aggregation and analysis, and situational assessment visualization of the power information network.

## Power Information Network Security Situation Quantification Method

In order to concretely reflect the state of the network security posture, the information of the element indicators gathered in the respective networks is realized as a centralized interactive process. Combined with the PIN-NSSQ architecture, different network levels are integrated to reflect the network application system complexity, network security equipment configuration defects, and other attribute characteristics of the information network (Hou et al., 2020; Qu et al., 2018). The three-dimensional security situational awareness element index system of the power information network is constructed from three dimensions of network reliability, threat, and vulnerability, and the mathematical characterization of element indexes is carried out.

### Cyber Security Situation Space Elements Index System

By constructing the spatial index system of network security posture elements as shown in Figure 1, it enables the coupling and interconnection of each element index and power information network application business under the cross of attribute dimensions. An accurate, systematic, and comprehensive representation of the security situation of the power information network can be realized, and the main indicators are as follows:

1) Network vulnerability indicators: when network vulnerability element index information is obtained, the statistics of vulnerabilities are targeted at all online devices, meanwhile increasing the integrity of the device configuration and the redundancy of the network topology.
2) Network reliability index: for the actual acquisition of data information in the power information network, a comprehensive analysis of redundancy and stability is conducted.

3) Network threat indicators: when the network is attacked, it will have an impact on the operation of the equipment in the network, using threatening element attributes to fully demonstrate the severity of the threat to the network.

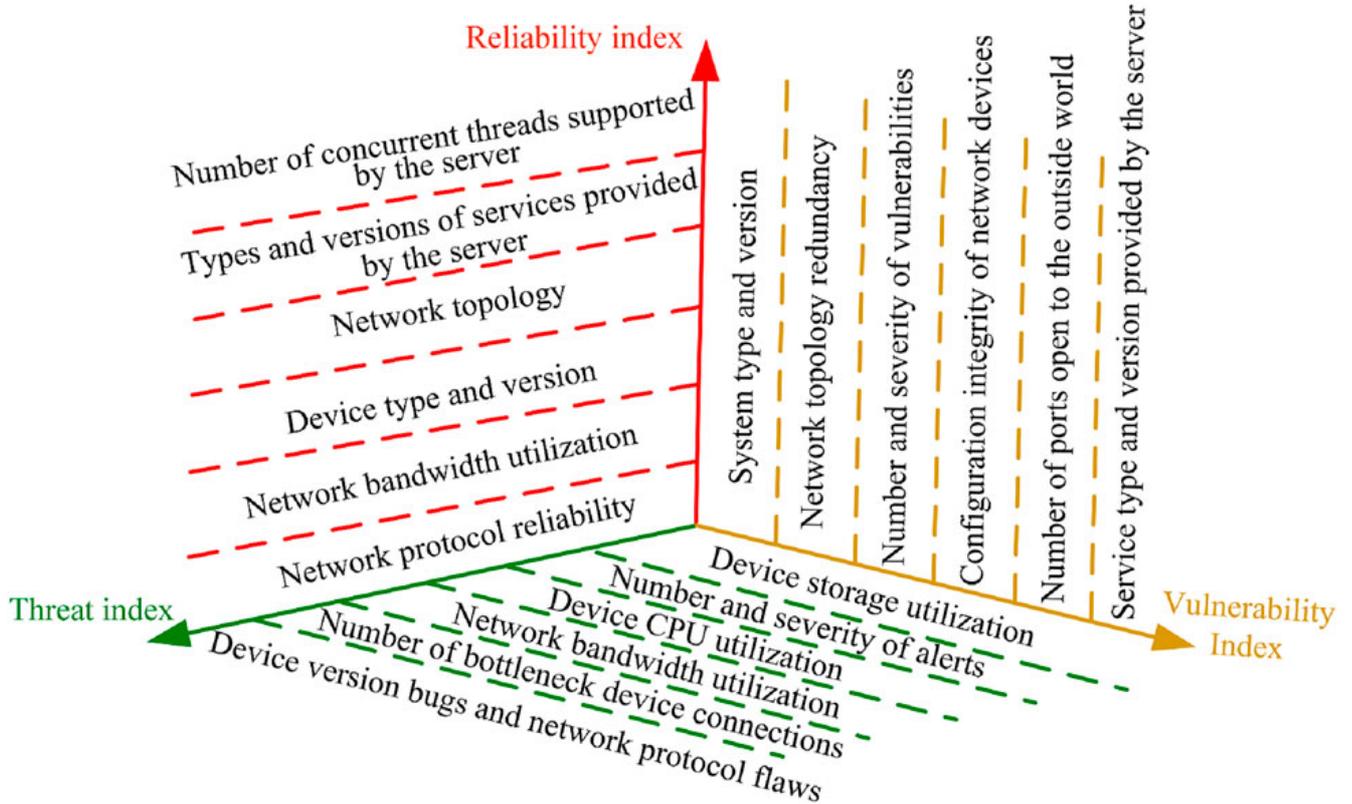

FIGURE 1. Spatial index system of network security situation elements.

## Mathematical Characterization of the Power Information Network Security Situation Elements

Mathematical characterization of the vulnerability index, reliability index, and threat index of the power information network, the calculation of elemental index values, and the weighted optimization of some parameters in combination with the actual network performance are realized. The foundation is laid for the quantification of power information network situation (Xie, 2022; Shahsavari et al., 2019).

Power information network vulnerability index $TR_{hk}$: it characterizes the cybersecurity vulnerability situation reflected by a system when different services on a system are subjected to cyberattacks at a certain time. The value of the vulnerability index for the system $HK$ at a given moment $t$ is as follows:

$$TR_{hk}(t) = \sum_{j=1}^{m} (v_j \times Rs_j(t)), \qquad (1)$$

where $m$ represents the number of services opened to the public by this host. $Rs_j(t) \quad (j = 1,2,3,\cdots,m)$ denotes the corresponding service situation value of the system $H_K$ at moment $t$. $v_j$ represents the weight of $S_j$, which reflects how many users the service hosts and how often it is used. The $v_j = SI_j / \sum_{t=1}^{m} SI_t$ is obtained using normalization of the service importance $SI_j$.

Power information network reliability index $Rs_j$: it characterizes the state of the network security reliability situation reflected by a service over a certain period of time when that service is subjected to a certain number of attacks. During the specified time period, define the moment $t$ as the index $Rs_j$ when this service $S_j$ is subjected to an attack.

$$Rs_j(t) = \sum_{i=1}^{n} (C_{ji} \times 100^{D_{ji}-1}), \qquad (2)$$

where $D_{ji}$ denotes the severity of the attack occurring at the moment $t$. $C_{ji}$ represents the frequency of attacks suffered at that moment. The number of types of attacks that occurred during the time period $t + \Delta t$ is denoted by $n$. $\Delta t$ is the attack time window determined by the network manager based on the attack alarm log data over a period of time. The higher the $Rs_j$ -value, the more severe the attack on service $S_j$ and the more and more urgent the reliability protection it needs. Administrators should quickly identify the appropriate solution.

Power information network threat index $RL$: it characterizes the threat situation reflected by the information network when the equipment or system service in the network is attacked. The calculation formula of the security threat value $R_L$ of the information network at time $t$ is as follows:

$$R_{Lq}(t) = \sum_{k=1}^{u} (w_k \times R_{Hk}(t)), \qquad (3)$$

where u is the number of active network devices, servers, and hosts in the network. $RH_k(t)$ is the improved network vulnerability index value, calculated by $RH_k(t) = \eta_k * TRH_k(t)$. $\eta$ is a correction factor determined according to the performance parameters at time $t$. $w_k$ represents the importance of the weight of active devices in the network, and the calculation method is $w_k = HI_k / \sum_{t=1}^{u} HI_t$. The higher the R value, the greater the security threat.

## Quantitative Model of Security Situation of the Power Information Network Based on the Evolutionary Neural Network

The quantification method of each dimensional element in the spatial index system is the core of the index system (Li et al., 2019). The quantitative model of perception factor index proposed in this article calculates the vulnerability index $TR_{hk}$ TRhk, reliability index $Rs_j$ Rsj, and threat index $R_L$ RL, respectively, and then comprehensively obtains the network security situation. In order to ensure the accuracy of the network security situation value and the efficient calculation of the process, we integrate the BP neural network optimized by the genetic evolutionary algorithm into the calculation process (Pizzuti, 2017). A quantitative model of security situation of the power information network based on the evolutionary neural network is constructed.

The model is mainly divided into two steps: BP neural network index hierarchical quantification and genetic algorithm security situation value optimization. The execution process is shown in Figure 2. According to the characteristics of the three-dimensional element index system, a suitable BP neural network is constructed (Liu S. et al., 2020). The massive historical data of the power information network security situation awareness database is used as the training data of the BP neural network. The trained BP neural network can output by the prediction function according to the basic data of the current situation of the information network. In the process of optimizing the security situation value of the genetic algorithm, the prediction result of the BP neural network integrated with the performance parameter adjustment is taken as the individual fitness value. Through selection, crossover, and mutation operations, the optimal security situation value under network security threats is found. The security situation of the power information network can be accurately quantified.

figure 2

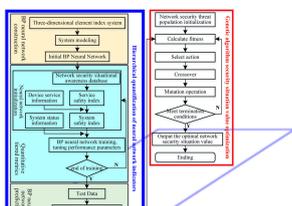

FIGURE 2. Quantitative model of the power information network security situation.

The specific steps are as follows:

**Step 1.** By collecting the original data of the dimension of network security situation elements, the key data is filtered and correlated and fused. The number and severity of attacks suffered by network services are analyzed. According to the three-dimensional element index system of the power information network, the number of nodes in the input layer $n$ n, the number of nodes in the hidden layer $l$ l, and the number of nodes in the output layer $m$ m are determined, and a suitable BP neural network is initially constructed.

**Step 2.** Calculate the network vulnerability index $TR_{hk}$ TRhk and network reliability index $Rs_j$ Rsj of the attack at time $t$ t by using the mathematical representation function, and the connection weights $\omega_{ij}$ ωij and $\omega_{jk}$ ωjk between the input layer, hidden layer, and output layer neurons are initialized. At the same time, the hidden layer threshold $a$ a and the output layer threshold $b$ b are initialized according to the device service information and system state information.

**Step 3.** Determine the optimal learning rate and activation function for the model (Sang, 2021). $R_{hk}(t)$ Rhk(t), $R_{sj}$ Rsj, $TR_{hk}$ TRhk, etc. are calculated by the performance parameter correction function and are defined as the input variable $X$ X. The hidden neuron solution is shown in Eq. 4.

$$H_j = f\left(\sum_{i=1}^{n} \omega_j x_i \cdots a_j\right), \qquad (4) \qquad Hj=f(\sum i=1n\omega ijxi\cdots aj),(4)$$

where $f(\cdot)$ f(·) is the hidden neuron activation function, and $f(x) = \frac{1}{1+e^{-x}}$ f(x)=11+e−x is adopted in this article.

**Step 4.** Output layer according to the hidden neuron $H$ H, the connection weight $\omega_{jk}$ ωjk and the threshold $b$ b and the quantitative prediction value $O$ O of the model situation is calculated.

$$O_k = \sum_{j=1}^{l} H_j \omega_{jk} - b_k. \qquad (5) \quad Ok=\sum j=1lHj\omega jk−bk.(5)$$

**Step 5.** Solve the individual fitness value $F$ Faccording to the situational quantitative prediction output $O$ O and the situational quantitative actual value $Y$ Y, the absolute value of the summation error is recorded as $E$ E. The specific solution method is as follows.

$$F = k\left(\sum_{i=1}^{n} abs(y_i - o_i)\right), \qquad (6) \qquad F=k(\sum i=1nabs(yi−oi)),(6)$$

where $n$ n is the number of output neurons in the quantized model. $y_i$ yi is the actual value of the situational quantification of the i-th neuron. $o_i$ oi is the model predicted output for the corresponding neuron. k is the coefficient, $k = 1,2,\cdots, m$ k=1,2,···,m.

**Step 6.** In terms of genetic selection, the calculation method of the selection probability $p_i$ pi of the quantitative value $i$ i of each time series sample situation is as follows.

$$p_i = \frac{f_i}{\sum_{j=1}^{N} f_j}. \qquad (7) \, pi=fi\sum j=1Nfj.(7)$$

In Eq. 7, $f_i = k/F_i$ fi=k/Fi, $F_i$ Fi is the fitness value of the situation quantification value $i$ i. The smaller the value, the better, so the inverse of the fitness value is calculated before the optimal situation quantification value is selected. $N$ N is the number of individuals in the population.

**Step 7.** Screen individuals through genetic selection, crossover, and mutation, so that individuals with accurate network security situation value predictions are retained, and individuals with abnormal situation values are eliminated. The new group not only integrates the information of the previous generation but also is better than the previous generation, so it iterates and loops until the termination condition is satisfied. The security situation value $RL$ under the threat of information network security is optimized, so as to accurately perceive the security situation of the entire power information network.

## Experiment Analysis of Examples

In order to verify the practicability and effectiveness of the method for quantifying the security situation of the power information network proposed in this article, the information and communication network of the power sector in a certain region is taken as an example. The experimental network topology is built through OpenNet, and three business data flows of mail service, marketing business, and conference phone system in the power information system are simulated.

## Comprehensive Calculation of Network Security Situation Element Index Values

Select a time period of 10 min, and the interval of each time period is 1 min, which is represented by $T_i \ (i = 1, 2, 3, \cdots, 10)$. Every time period, hacking tools are used to launch attacks on the power communication network, including server scans on the internal network, Ping of death, DoS, and other conventional network attacks.

Mathematically characterizing the attack severity of different services of each system in each period, and the spatial index system can clearly reflect the network status of a certain type of power application business under attack. Taking the power marketing business as an example, Figure 3 intuitively shows that 16 different states of the power information network are randomly selected by simulating different attack methods in 10 consecutive minutes.

figure 3

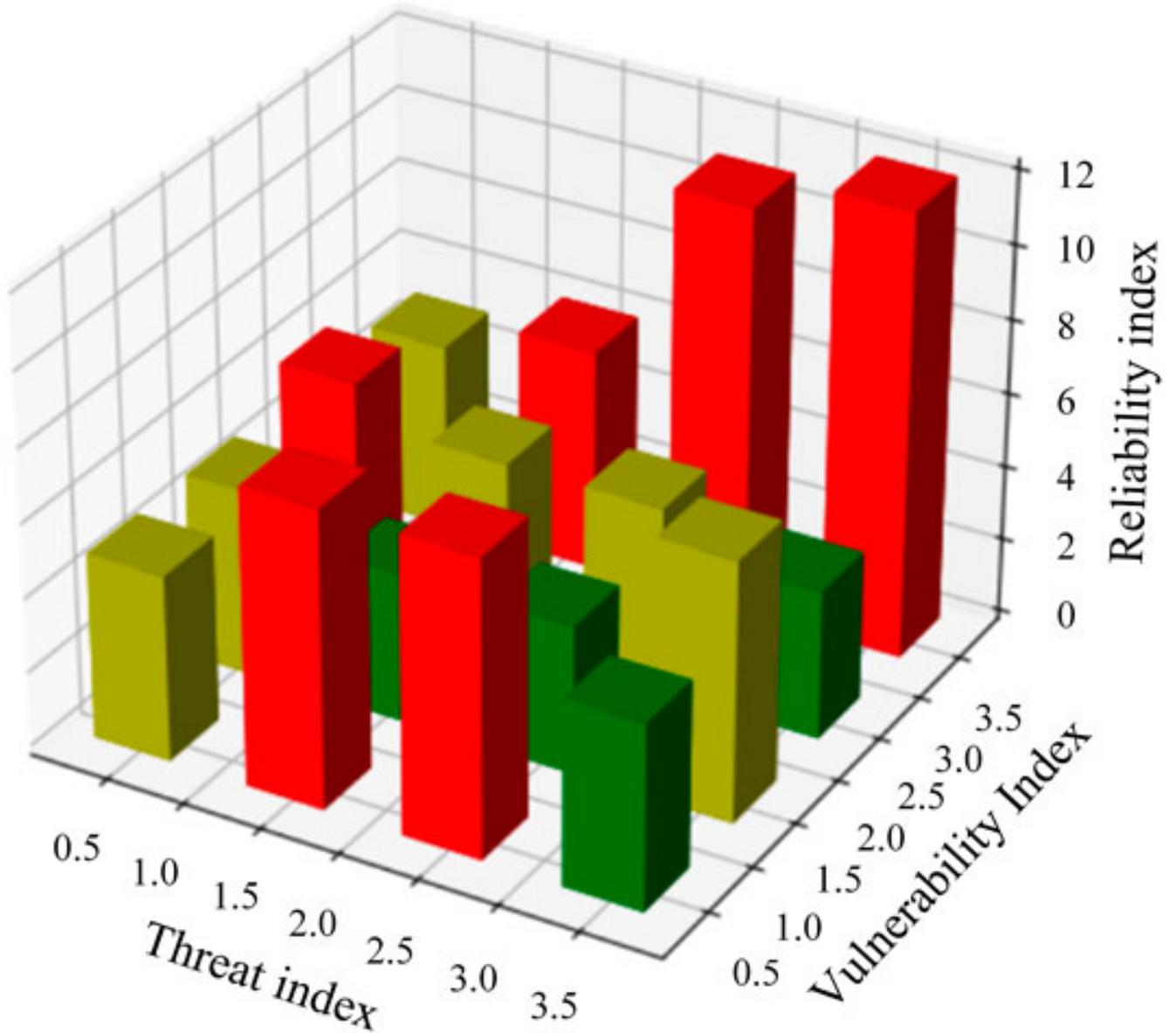

**FIGURE 3.** Spatial element index simulation under the attack state.

As shown in Figure 3, the composite index of vulnerability index is 3.5, and the composite index of threat index is 3.5. When both indices are high, it indicates that there are more vulnerabilities in the servers and other equipment running the power marketing business. The attack suffered is highly targeted to this type of business. At this time, the comprehensive index of the reliability index corresponding to the spatial index system is 11.8, and the communication network urgently needs a high-reliability solution for security protection. However, when the composite index of threat indicators is higher, it also reaches 3.5. The composite index of network equipment or application system vulnerability index is only about 0.5. The comprehensive index of reliability index only reaches level 3, indicating that the entire information and communication network does not need maintenance of reliability programs. The marketing business system may be able to directly detect such attacks, or the device has strong self-healing ability.

Therefore, after analyzing the simulation results of the network security situation element space index system, the application business characteristics should be combined when quantifying the security situation value of the power information network. According to the different dimension indices obtained, the security situation of the power information network can be evaluated more accurately, systematically, and comprehensively.

## Validation of the Quantification Method for Security Situation of the Power Information Network

In the event of a network attack, it is analyzed based on the relevant performance information and log data of the firewall and server host. Take 4,000 sets of data, randomly select 3,900 sets of data to train BP neural network, and 100 sets of data as test data. The fitting performance of the BP neural network is tested by algorithm comparison.

The output of the proposed power information network security situational awareness prediction algorithm is compared with the actual security situation value, and compared with the newer research literature (Wang et al., 2020), the network security situation quantification method is based on the improved LSTM neural network. The results are shown in Figure 4A. The prediction error of the network security situation quantification value of the algorithm in this article is shown in Figure 4B.

figure 4

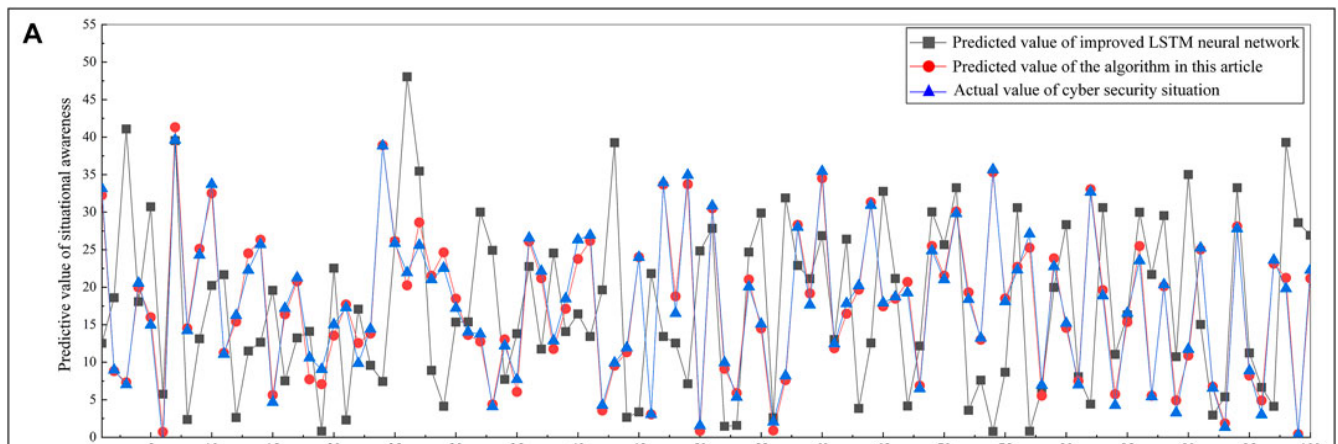

**FIGURE 4.** Comparison experiment of network security situation quantification results: (A) Comparison between the predicted value and the actual value. (B) Prediction error.

It can be seen from Figure 4A that the model in this article can accurately predict the situation value of network equipment, servers, and hosts running in the power information network when they are under attack threats, and compared with the improved LSTM neural network, the predicted value can be closer to the actual value. Figure 4B shows that the error between the predicted output of the network model and the expected output is mostly maintained between [−0.02, 0.02], which can effectively calculate the individual fitness. It lays a good foundation for the optimization of the security situation value of the comprehensive power information network by using the genetic evolution algorithm.

## Conclusion

Based on the background of the new power information network, this article analyzes the specific characteristics of the information network and the power services it carries. An evolutionary neural network security situation quantification method for the power information network is proposed. The proposal of this technology improves the quantification accuracy of the security situation of the power information network to a certain extent. The experimental simulation results verify the following main conclusions:

1) A PIN-NSSQ architecture of a power information network is designed, taking into account the importance of network performance and power services.
2) From the three dimensions of reliability, threat, and vulnerability, a three-dimensional element index system of network security situation is established. At the same time, the mathematical representation of the element indicators around the comprehensive indicators of indicators is proposed, which lays a core foundation for the construction of a quantitative model of network security situation.
3) A new method to quantify the security situation of the power information network is proposed. The algorithm adopts BP neural network data fitting, and introduces the genetic evolution algorithm to realize the accurate calculation of the network security situation value. The method provides powerful data decision support for power network administrators to master and strengthen network security.

## Data Availability Statement

The original contributions presented in the study are included in the article/Supplementary Material; further inquiries can be directed to the corresponding author.

## Author Contributions

QY: designed this study. YP: contributed to the power information network security posture quantification system architecture. LK: contributed to the power information network security situation quantification method. FZ: collected and cleansed the data. BY: conducted simulation experimental analysis based on actual power information topologies. All authors contributed to the writing of the article and all agreed to the submitted version of the article.

## Funding

This research is supported by the Science and Technology Development Plan Project of Jilin Province (grant number 20210201049GX), the Science and Technology Projects of Education Department of Jilin Province (grant numbers: JJKH20191262KJ and JJKH20191258KJ).

## Conflict of Interest

The authors declare that the research was conducted in the absence of any commercial or financial relationships that could be construed as a potential conflict of interest.

## Publisher's Note

All claims expressed in this article are solely those of the authors and do not necessarily represent those of their affiliated organizations, or those of the publisher, the editors, and the reviewers. Any product that may be evaluated in this article, or claim that may be made by its manufacturer, is not guaranteed or endorsed by the publisher.

CrossRef Full Text  |  Google Scholar

Edited by:

Chengzong Pang,Wichita State University, United States

Reviewed by:

Cheng Qian,First Affiliated Hospital of Zhengzhou University, China
Jun Yin, North China University of Water Resources and Electric Power, China

**\*Correspondence:** Lei Kou, koulei1991@hotmail.com

**Disclaimer:** All claims expressed in this article are solely those of the authors and do not necessarily represent those of their affiliated organizations, or those of the publisher, the editors and the reviewers. Any product that may be evaluated in this article or claim that may be made by its manufacturer is not guaranteed or endorsed by the publisher.